\documentclass[10pt,twocolumn,letterpaper]{article}

\usepackage[pagenumbers]{cvpr}

\usepackage{amsmath,amssymb,amsfonts}
\usepackage{graphicx}
\usepackage{subcaption}
\usepackage{booktabs}
\usepackage{array}
\usepackage{xcolor,colortbl}

\definecolor{cvprblue}{rgb}{0.21,0.49,0.74}
\usepackage[breaklinks,colorlinks,allcolors=cvprblue]{hyperref}
\usepackage{amsfonts}
\usepackage{amsmath}
\usepackage{amssymb}
\usepackage{graphicx}
\usepackage{cuted}
\usepackage{caption}
\usepackage{subcaption}
\usepackage{placeins}
\usepackage{booktabs}
\usepackage{algorithm}
\usepackage{algpseudocode}

\newcommand{\inst}[1]{$^{#1}$}

\begin{document}

\title{SAM-V: Geometry-Aware Segment Anything for Multi-View Instance Segmentation} 

\author{%
  Jiangshan Gong\inst{1}$^\ast$ \qquad
  Yuqun Wu\inst{1}$^\ast$ \qquad
  Qiqian Fu\inst{1} \qquad
  Yao Xiao\inst{1} \\
  Chuhang Zou\inst{2}  \qquad
  Shenlong Wang\inst{1} \qquad
  Derek Hoiem\inst{1} \\[0.8em]
  $^{1}$University of Illinois at Urbana-Champaign \quad $^{2}$Meta
}

\twocolumn[{%
  \renewcommand\twocolumn[1][]{#1}%
  \maketitle
  \vspace{-1.5em}
  \begin{center}
    \captionsetup{type=figure}
    \includegraphics[width=0.75\textwidth]{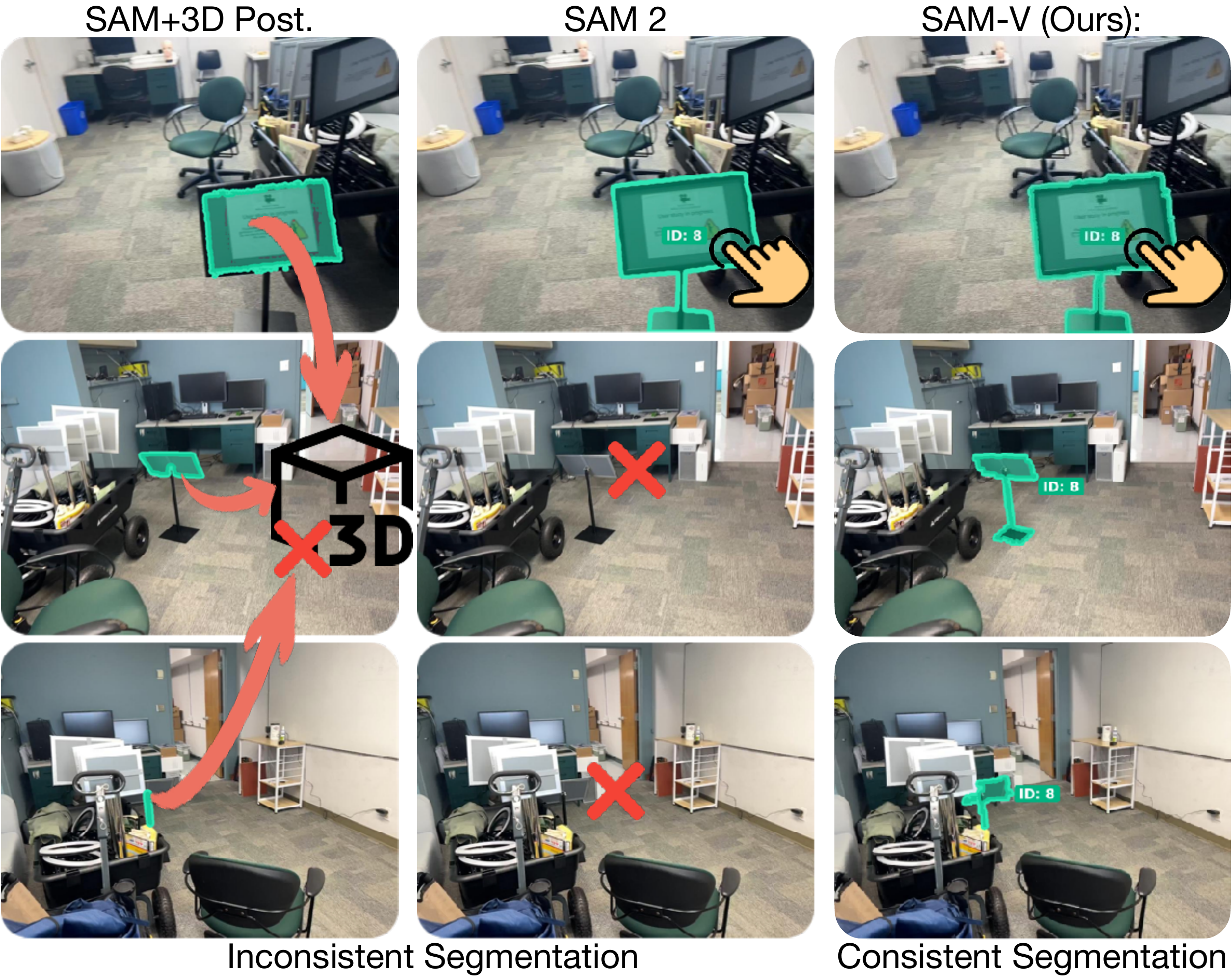}
    \captionof{figure}{
      \textbf{SAM-V enables consistent object segmentation across large viewpoint changes.} Directly applying SAM masks with 3D postprocessing for association can lead to inconsistent cross-view segmentation, while SAM2 may lose track of the target when the viewpoint changes significantly. In contrast, SAM-V conditions mask decoding on multi-view geometry, keeping the same target segmented across views.}
    \label{fig:teaser_figure}
  \end{center}
  \vspace{0.5em}
}]

\renewcommand{\thefootnote}{\ensuremath{\ast}}
\footnotetext{Equal contribution}
\renewcommand{\thefootnote}{\arabic{footnote}}

\begin{abstract}
Consistent multi-view object segmentation is critical for 3D perception and robotics, yet remains challenging under severe viewpoint and occlusion changes. Existing methods typically perform 3D instance segmentation on point clouds or rely on offline 2D mask-matching pipelines. However, 3D instance segmentation is limited by scarce 3D annotations, while offline 2D matching suffers from object identity ambiguity across frames. To leverage strong 2D and 3D priors jointly, we propose SAM-V (Geometry-Aware \textbf{S}egment \textbf{A}nything for \textbf{M}ulti-\textbf{V}iew Instance Segmentation). Instead of combining the two priors through post-hoc matching, SAM-V directly integrates features from a feed-forward geometry model (VGGT) into a 2D segmentation foundation model (SAM), trained end-to-end for cross-view instance prediction. SAM-V introduces a prompt-fusion mechanism that enriches sparse SAM prompt tokens with view-specific camera tokens and local VGGT features, making the prompt representation both view-aware and spatially grounded, together with a mask decoder that attends to dense 2D and 3D features. By conditioning the mask decoding directly on multi-view geometry, SAM-V produces consistent multi-view segmentation of a prompted object in a single forward pass without offline mask matching or explicit 3D reconstruction. On the IGGT 3D tracking benchmark, where consistent instance identity across frames directly determines performance, SAM-V improves overall IoU by 5 points and frame-level recall by 12 points on the ScanNet++ split over the state-of-the-art multi-view instance segmentation baseline and leads on all metrics in the zero-shot ScanNet split. Our code and pretrained models are available in the
\href{https://github.com/gong208/SAM-V.git}{project repository}.
\end{abstract}

\section{Introduction}

While single-image segmentation has seen immense progress through large-scale visual foundation models~\cite{Kirillov2023SegmentA, Cheng2021MaskedattentionMT, He2017MaskR}, consistent multi-view object segmentation remains a bottleneck for 3D perception and robotic applications~\cite{Schnberger2016StructurefromMotionR, mildenhall2020nerf}. 
For a robot to perform real-world tasks, e.g., grasping a target object or tidying a room, it must reliably recognize the same physical object across views despite severe viewpoint, occlusion, and scale variations. 

Existing multi-view and 3D segmentation methods typically fall into two paradigms. The first operates directly within explicit 3D representations such as point clouds~\cite{Jain2024ODINAS,Jiang2020PointGroupDP,Kolodiazhnyi2023TopDownBB,Schult2022Mask3DMT,Jeong2026MVSAMMP}. These methods depend heavily on reconstruction quality and are limited by scarce 3D annotations. 
The second relies on offline projection-and-matching pipelines to associate per-view 2D masks~\cite{Guo2023SAMguidedGC,Yang2023SAM3DSA,Xu2023SAMPro3DLS}, which suffer from object identity ambiguity across frames. Recent feed-forward approaches such as IGGT~\cite{Li2025IGGTIG} avoid both explicit reconstruction and offline association by learning instance-aware geometry features end-to-end, but they do not leverage the segmentation priors of 2D foundation models, making cross-view mask prediction hard.
Integrating a strong 2D segmentation prior with a multi-view geometry prior is thus an appealing alternative. However, fusing dense image features alone is insufficient. The sparse prompt specified in a single view remains view-agnostic, so the decoder cannot resolve which view and location the prompt refers to, and predictions drift across viewpoints (Sec.~\ref{sec:ablation}).
Achieving consistency requires grounding the prompt itself in the multi-view geometry.

In this work, we propose \textbf{SAM-V}, an end-to-end framework for prompt-conditioned multi-view instance segmentation. Rather than relying on post-hoc association, SAM-V conditions the segmentation process on multi-view geometry \textit{before} mask decoding. 
We couple SAM~\cite{Kirillov2023SegmentA} and VGGT~\cite{Wang2025VGGTVG} through a novel geometry-aware fusion mechanism at both the image and prompt levels. Crucially, we enrich sparse SAM prompt tokens with view-specific camera tokens and local VGGT features sampled at the prompted locations. This anchors the prompt to the multi-view context, making the representation both view-aware and spatially grounded.
By decoding from this unified representation, SAM-V produces consistent cross-view masks in a single forward pass, providing the lightweight 3D inference while bypassing the latency of offline merging. 

While SAM-V natively supports target-specific segmentation from sparse prompts, real-world autonomy often requires scene-level understanding. We therefore introduce an every-object inference procedure (Sec.~\ref{sec:every_object_inference}), enabling SAM-V to transition from prompt-conditioned tracking to segmenting all object instances consistently across a scene.

Empirically, we evaluate SAM-V on the ScanNet++ and ScanNet scenes of the IGGT 3D tracking benchmark~\cite{Li2025IGGTIG} with a unified protocol for cross-view consistent instance segmentation. Under this protocol the strongest baseline is PanSt3R~\cite{Zust2025PanSt3RMC}, over which SAM-V improves IoU by 5 points and frame-level recall by 12 points. These results indicate that geometry-aware fusion effectively adapts 2D and multi-view priors to multi-view mask prediction. 

In summary, our contributions are:
\begin{itemize}
    \item We propose SAM-V, an end-to-end framework that couples SAM with VGGT geometry for consistent multi-view instance segmentation in a single forward pass.
    \item We introduce a geometry-aware prompt fusion mechanism that grounds sparse prompts with camera tokens and local geometry features, clearly specifying input views.
    \item We extend SAM-V with an every-object inference procedure for scene-level segmentation. On the IGGT 3D tracking benchmark, it outperforms the strongest baseline PanSt3R by 5 O-IoU and 12 frame-level recall points on ScanNet++, and leads on all metrics in the zero-shot ScanNet split.    

\end{itemize}

\section{Related Work}
\subsection{2D Dense Segmentation}
Dense image segmentation has been widely studied in semantic, instance, and panoptic segmentation, with representative frameworks including Mask R-CNN~\cite{He2017MaskR}, MaskFormer~\cite{Cheng2021PerPixelCI}, Mask2Former~\cite{Cheng2021MaskedattentionMT}, and MaskDINO~\cite{Li2022MaskDT}. More recently, promptable foundation segmentation models such as SAM~\cite{Kirillov2023SegmentA} and its variants~\cite{Ke2023SegmentAI,Zou2023SegmentEE,Li2023SemanticSAMSA} have shown strong mask quality from sparse prompts, while SAM2~\cite{Ravi2024SAM2S} and video object segmentation methods~\cite{Cheng2021RethinkingSN,Cheng2022XMemLV,Yang2021AssociatingOW,Yang2022DecouplingFI} extend mask propagation to video settings. Most recently, SAM3~\cite{Carion2025SAM3S} generalizes the promptable paradigm from spatial prompts to concept prompts, detecting, segmenting, and tracking all instances matching a short noun phrase or image exemplar across images and videos.

These models provide strong priors for mask quality and promptability, and memory-based propagation maintains object identity reliably when observations vary smoothly. SAM2 remains a strong baseline in this regime, as shown in Table~\ref{tab:hypersim_overall_iou}. These models are designed and supervised around image appearance and temporal continuity rather than explicit multi-view 3D geometry. Associating the same physical object across widely separated viewpoints, or disambiguating visually similar instances observed in different views, therefore remains an open challenge for this family. Rather than propagating masks along a temporal axis, SAM-V builds on SAM and injects VGGT-derived multi-view geometric information to enable consistent object segmentation across views.

\begin{figure*}[ht]
    \centering
    \includegraphics[
        page=1,
        width=\linewidth,
        clip
    ]{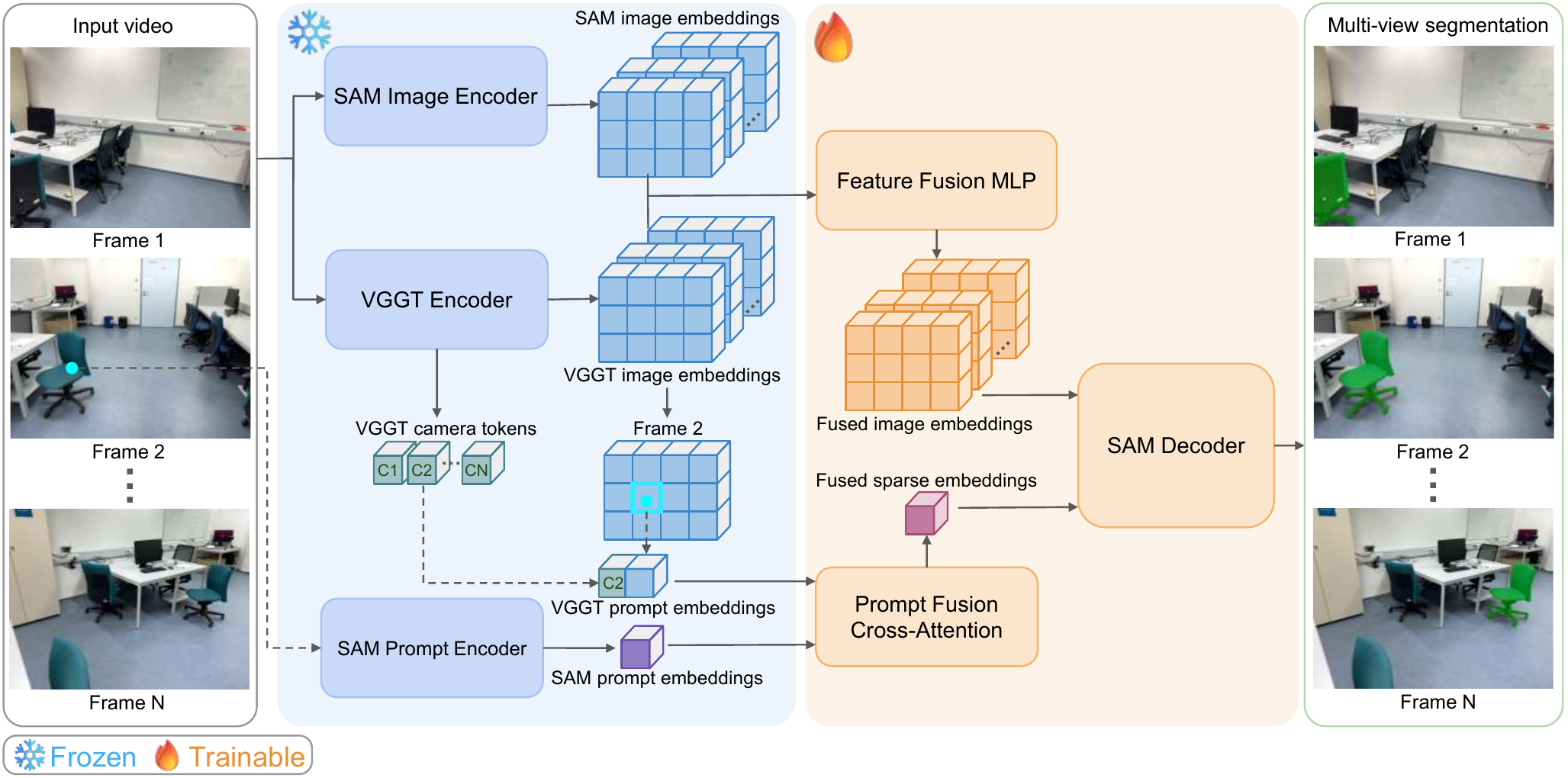}
    \caption{
        \textbf{Overview of the proposed SAM-V architecture.}
        Given $N$ frames from the same scene and a sparse point prompt on one view, SAM-V predicts masks for the same target object across all views. 
        The blue dot on Frame 2 indicates the user-provided point prompt. 
        SAM-V integrates SAM image and prompt embeddings with VGGT geometry-aware image features, camera tokens, and local prompt-location features, and decodes the fused representations with a finetuned SAM mask decoder.
        }
    \label{fig:model_architecture}
\end{figure*}

\subsection{3D and Multi-View Instance Segmentation}
Instance segmentation in 3D scenes is commonly performed directly on point clouds or reconstructed 3D representations~\cite{Jiang2020PointGroupDP, Schult2022Mask3DMT, Kolodiazhnyi2023TopDownBB, Jain2024ODINAS}, where the shared metric space itself provides cross-view association. Point-SAM~\cite{Zhou2024PointSAMP3} brings SAM-style promptability into this space, and concurrent work MV-SAM~\cite{Jeong2026MVSAMMP} grounds image prompts in an explicit pointmap representation. MV-SAM shares our observation that sparse prompts must be grounded in multi-view geometry, but resolves the prompt in a lifted 3D space, whereas SAM-V keeps both prompts and outputs in the original 2D views, reducing the dependence on reconstruction quality.

A second family uses SAM as a 2D mask generator and associates per-view masks through projection, 2D--3D lifting, graph optimization, or prompt alignment~\cite{Yang2023SAM3DSA, Xu2023SAMPro3DLS, Guo2023SAMguidedGC, Takmaz2023OpenMask3DO3, Nguyen2023Open3DISO3, Yin2023SAI3DSA}, establishing that 2D foundation priors transfer to multi-view settings without retraining. V\textsuperscript{2}-SAM~\cite{Pan2025V2SAMMS} likewise observes that spatially grounded prompts do not transfer across viewpoints, pairing SAM2 with multi-prompt experts for cross-view correspondence. Building on these insights, SAM-V incorporates multi-view geometry before mask generation rather than only linking independently produced masks afterward.

Recent feed-forward approaches jointly model geometry and instance semantics on reconstruction backbones~\cite{Wang2023DUSt3RG3, Cabon2025MUSt3RMN, Wang2025Continuous3P, Wang2025VGGTVG, Wang2025pi3PV, Keetha2025MapAnythingUF}: IGGT~\cite{Li2025IGGTIG} learns instance-grounded features for 3D tracking, extended to dynamic scenes by IGGT4D~\cite{Zou2026IGGT4DS4}, PanSt3R~\cite{Zust2025PanSt3RMC} predicts multi-view consistent panoptic segmentation, and FAST3DIS~\cite{Li2026FAST3DISFA} replaces post-hoc clustering with learned 3D anchors. 
Closest to our setting is VGGT-Segmentor~\cite{Gao2026VGGTSegmentorGC}, which builds a custom segmentation head on VGGT features to find correspondences between egocentric and exocentric views, and observes pixel-level projection drift when prompts are not anchored to a specific image location, an observation our ablation aligns with (Sec.~\ref{sec:ablation}). To address this drift without training a task-specific head, SAM-V instead injects multi-view geometry into both the dense image features and the sparse prompt tokens of a SAM-style decoder, inheriting SAM's 2D priors while keeping prompts spatially anchored.

\section{Method}
SAM-V aims to predict instance masks given multi-view images (Fig.~\ref{fig:model_architecture}). 
We first introduce multi-view instance segmentation with prompts, and then extend it to every-object segmentation. 

\noindent \textbf{Task Definition}:
Given $N$ RGB images $\mathcal{I}=\{I_1,\ldots,I_N\}$, the goal is to segment the same object in every view where it appears.
The target is specified by point prompts $\mathcal{P}=\{(u_i,v_i,n_i)\}_{i=1}^{N_p}$, where $(u_i,v_i)$ is the 2D coordinate of the $i$-th prompt and $n_i\in\{1,\ldots,N\}$ is its frame index, and prompts may come from one or multiple views.
The output is frame-wise masks $\hat{\mathcal{M}}=\{\hat{M}_1,\ldots,\hat{M}_N\}$, where $\hat{M}_n \in [0,1]^{H_0 \times W_0}$. If the target is not visible in frame $n$, $\hat{M}_n$ should be the all-zero mask.

\noindent \textbf{Overall Approach}: 
2D segmentation models provide strong per-view instance priors, while feed-forward geometry models  provide multi-view consistent spatial information. To exploit both priors, SAM-V fuses the two feature streams inside the model before mask decoding, rather than combining their outputs post hoc. The fusion operates at two levels: dense image features from both encoders are merged into a unified representation, and the input prompts are enriched with 2D positional and 3D geometric context so that they unambiguously specify the target object across views. The mask decoder then attends to the fused features under these geometry-grounded prompts to predict a mask for the same instance in every view.

\subsection{Instance and Geometry Feature Integration}
\label{sec:feature_integration}

Given the $N$ input frames, we extract image embeddings from the frozen SAM and VGGT encoders,
$\mathbf{F}_{\text{SAM}} \in \mathbb{R}^{N \times H \times W \times C_{\text{SAM}}}$ and
$\mathbf{F}_{\text{VGGT}} \in \mathbb{R}^{N \times H \times W \times C_{\text{VGGT}}}$,
where $H \times W$ is the feature resolution.
SAM features provide strong 2D segmentation cues, while VGGT features encode multi-view consistent geometry. We concatenate them along the channel dimension and project the result back to the SAM embedding dimension $\mathbb{R}^{N \times H \times W \times C_{\text{SAM}}}$:
\[
\mathbf{F}_{\text{fused}}
=
\operatorname{MLP}
\left(
\operatorname{Concat}
(\mathbf{F}_{\text{SAM}}, \mathbf{F}_{\text{VGGT}})
\right).
\]

The fused features of all $N$ views are flattened into a single token sequence, so the mask decoder attends over every view jointly. Positional encodings are adjusted accordingly so that each prompt aligns with its prompted view.

\subsection{Geometry-Aware Prompt Inputs}
\label{sec:geometry_aware_prompt_inputs}

In the original SAM setting, prompts are interpreted in a single image feature map. In our setting, the decoder attends over $\mathbf{F}_{\text{fused}}$ spanning $N$ frames, so each prompt must convey both the prompted view and the queried location. We therefore augment the SAM prompt embeddings with VGGT-derived view-level and point-level context.

Given the prompt points $\mathcal{P}=\{(u_i,v_i,n_i)\}_{i=1}^{N_p}$, we first obtain SAM prompt embeddings
$\mathbf{P}_{\text{SAM}} = E_{\text{prompt}}(\mathcal{P})$
from the frozen SAM prompt encoder.
For each prompt $i$, we gather two pieces of geometric context.
First, the camera token of the prompted frame,
$\mathbf{G}_{\text{cam}}[i] = \mathbf{T}_{\text{cam}}[n_i]$,
where $\mathbf{T}_{\text{cam}} \in \mathbb{R}^{N \times C_{\text{cam}}}$ are the frame-level camera tokens.
Second, the local geometry feature
$\mathbf{G}_{\text{local}}[i]$, taken from the VGGT patch containing $(u_i, v_i)$ in frame $n_i$. The precise location is already encoded by the SAM prompt embedding, so patch-level context suffices.
We concatenate them into the prompt context
$\mathbf{G}_{\text{prompt}} = \operatorname{Concat}(\mathbf{G}_{\text{cam}}, \mathbf{G}_{\text{local}})$,
and fuse it with the SAM prompt embeddings using cross-attention:
\[
\mathbf{P}_{\text{fused}}
= \operatorname{CrossAttn}(Q{=}\mathbf{P}_{\text{SAM}},\ K{=}\mathbf{G}_{\text{prompt}},\ V{=}\mathbf{G}_{\text{prompt}}).
\]

The camera token indicates which view the prompt comes from, while the local VGGT feature anchors the spatial information at the queried location, indicating where the prompt lies in the scene. Together they give the decoder a view-aware and spatially grounded prompt representation. We validate the contribution of each component in Sec.~\ref{sec:ablation}.

The mask decoder then takes the fused features and prompts and predicts the multi-view masks in a single pass:
\[
\hat{\mathcal{M}}
= \operatorname{Decoder}\bigl(\mathbf{F}_{\text{fused}},\, \mathbf{P}_{\text{fused}}\bigr),
\]
where the decoder attends over the tokens of all $N$ views jointly, and the output logits are split by view to form $\hat{\mathcal{M}}$.
 
\subsection{Training Objective}
\label{sec:training_objective}

During training, the SAM image encoder, VGGT encoder, and SAM prompt encoder are kept frozen. We optimize the feature-integration MLP, the geometry-aware prompt-fusion module, and the SAM mask decoder, which is initialized from pretrained SAM weights.

SAM-V is trained in two stages: large-scale pretraining on the synthetic Hypersim dataset, whose rendered ground truth provides exact multi-view instance masks, followed by finetuning on ScanNet++ to adapt to real-world capture, where annotations are typically noisier.
Following SAM~\cite{Kirillov2023SegmentA}, we supervise mask prediction with a weighted combination of focal loss~\cite{Lin2017FocalLF} and Dice loss~\cite{Milletar2016VNetFC}:
$\mathcal{L}_{\mathrm{mask}} = 20\mathcal{L}_{\mathrm{focal}} + \mathcal{L}_{\mathrm{dice}}$
, where the focal loss addresses foreground--background imbalance and the Dice loss encourages overlap with the ground-truth masks.
In the finetuning on ScanNet++, we add a mask-confidence loss $\mathcal{L}_{\mathrm{IoU}}$, the mean squared error between the decoder's predicted IoU and the actual IoU of the predicted mask against the ground truth, so that candidate masks can be reliably ranked during every-object inference.
Additional implementation details are provided in Sec.~\ref{exp:details} and the supplement.

\begin{figure*}[t]
    \centering
    \includegraphics[width=\linewidth]{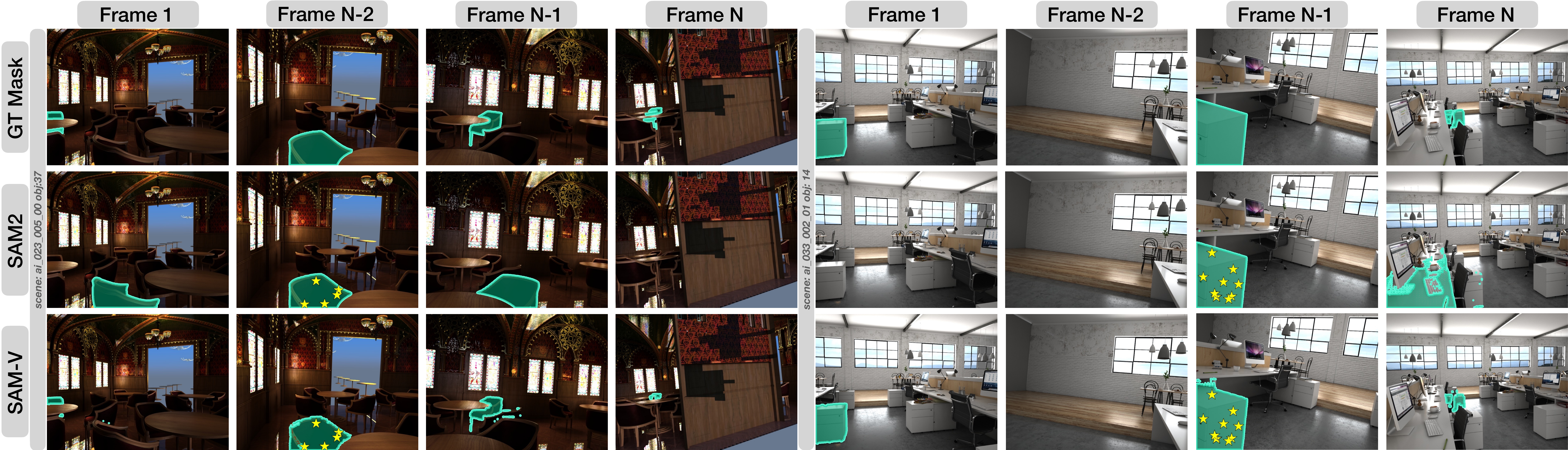}
    \caption{
    \textbf{Comparison with SAM2 on Hypersim.} Yellow stars denote prompt points and green masks the predicted objects. SAM-V produces more accurate and consistent masks than SAM2 in scenes with visually similar objects and under large viewpoint changes.
    }
    \label{fig:qual_sam2_vggtsam}
\end{figure*}

\subsection{Every-Object Inference}
\label{sec:every_object_inference}

To extend SAM-V to every-object inference, we use SAM proposals to generate prompts automatically.
Given $N$ input frames, we first run SAM~\cite{Kirillov2023SegmentA} on each frame independently to obtain 2D proposal masks. For each proposal, we sample point prompts inside its mask region and feed this prompt group together with all $N$ frames into SAM-V to predict a candidate multi-view mask.

Proposals from different frames or overlapping regions may correspond to the same physical object, producing duplicate candidates. However, because each candidate is already a multi-view mask, duplicates of the same object largely coincide across all views. We therefore apply non-maximum suppression based on mask overlap to remove redundant predictions. The remaining masks form the final every-object multi-view segmentation output. Details are given in the supplementary material.

\section{Experiment}

We evaluate SAM-V to answer three questions. 

\textbf{Q1}: Can the model segment the same prompted object consistently across multiple views, including views where the object undergoes large viewpoint changes or is absent? 

\textbf{Q2}: Can the prompt-conditioned model be extended to every-object scene-level inference and remain competitive with methods designed for 3D tracking? 

\textbf{Q3}: How important is the local VGGT feature in grounding sparse prompts to the correct image location? 

To answer these questions, we conduct controlled prompt-conditioned evaluation on Hypersim, every-object evaluation on the ScanNet++ and ScanNet splits from the IGGT 3D tracking benchmark, and an ablation study for the geometry-aware prompt-fusion design.

\subsection{Experiment Setup}
\label{exp:details}
\noindent \textbf{Training Details.}
We train SAM-V on Hypersim~\cite{Roberts2020HypersimAP} and ScanNet++~\cite{Yeshwanth2023ScanNetAH}, which provide multi-view RGB images with frame-consistent instance annotations, using the objective of Sec.~\ref{sec:training_objective}.
Each training sample contains $8$ frames of one scene for one target instance: $4$ positive frames where the target is visible and $4$ negative frames where it is absent, so the model is explicitly supervised to predict all-zero masks in views without the target.
On Hypersim, we employ a progressive curriculum that gradually increases viewpoint diversity among positive frames and shifts from multi-frame to single-frame prompting, letting the model first learn prompt grounding in easier configurations before tackling single-frame prompts under pose-diverse views. We then fine-tune on ScanNet++ with pose-diverse frames and $3$--$5$ prompt points.
Full curriculum schedules, sampling thresholds, and hyperparameters are provided in the supplement.

\noindent \textbf{Evaluation Metrics.}
We use three metrics that probe complementary failure modes.
\textbf{T-mIoU}~\cite{Li2025IGGTIG} averages the per-frame IoU over frames where the ground-truth object is visible, measuring per-view segmentation quality.
\textbf{O-IoU} concatenates all frame-wise masks into one multi-view mask and computes a single overall IoU, measuring aggregate quality over the full sequence, including frames where the object is absent.
Finally, frame-level \textbf{precision and recall} at IoU thresholds from $0.1$ to $0.9$ capture detection behavior: recall reflects how completely the object is recovered across views, while precision penalizes spurious predictions. A prediction counts as a true positive if its IoU with the ground-truth mask exceeds the threshold; a non-empty prediction without a match is a false positive, and an unmatched non-empty ground-truth mask is a false negative. P@50/R@50 denote threshold $0.5$. Formal definitions are given in the supplement.

\noindent \textbf{Baselines}: We compare against video tracking method SAM2~\cite{Ravi2024SAM2S}, SAM + 3D post-processing method Point-SAM~\cite{Zhou2024PointSAMP3}, 3D instance segmentation method ODIN~\cite{Jain2024ODINAS}, instance-aware feature learning method IGGT~\cite{Li2025IGGTIG}, and geometry feature + learnable query method PanSt3R~\cite{Zust2025PanSt3RMC}. All methods operate from RGB inputs alone, with no sensor depth or ground-truth poses. ODIN and Point-SAM require 3D inputs, so we provide both with the pointmaps from VGGT~\cite{Wang2025VGGTVG}. Note that PanSt3R and ODIN discover instances through learnable scene-level queries and cannot accept user prompts, whereas SAM2, Point-SAM, and SAM-V are promptable. See full details in the supplement.

\subsection{Comparison with Baselines}
\label{sec:baselines}
We evaluate SAM-V on Hypersim~\cite{Roberts2020HypersimAP}, ScanNet++~\cite{Yeshwanth2023ScanNetAH}, and ScanNet~\cite{Dai2017ScanNetR3}. 
On Hypersim, we construct an evaluation set for \textbf{multi-view single instance segmentation} from held-out scenes disjoint from training, selecting instances visible in at least $8$ frames above the minimum area threshold, yielding $1{,}788$ evaluated instances.
On ScanNet++ and ScanNet, we follow the IGGT 3D tracking benchmark~\cite{Li2025IGGTIG}, using $58$ ScanNet++ objects and $66$ ScanNet objects for \textbf{every object segmentation} evaluation.

\begin{table}[t]
\centering
\caption{\textbf{Comparison with SAM2 on the Hypersim test split under two multi-view frame sampling strategies.} Continuous denotes the setting where input frames are sampled from nearby camera poses, while diverse denotes the setting where positive frames are selected to have larger viewpoint variation.
}
\label{tab:hypersim_overall_iou}
\begin{tabular}{lcccc}
\toprule
& \multicolumn{2}{c}{\textbf{Continuous}} 
& \multicolumn{2}{c}{\textbf{Diverse}} \\
\cmidrule(lr){2-3} \cmidrule(lr){4-5}
Method & O-IoU & R@50 & O-IoU & R@50 \\
\midrule
SAM2  & 63.5 & \textbf{76.8} & 48.1 & 59.6 \\
SAM-V & \textbf{64.2} & 75.5 & \textbf{62.4} & \textbf{71.9} \\
\bottomrule
\end{tabular}
\vspace{-1.5em}
\end{table}

\paragraph{Multi-view single instance segmentation.}
On Hypersim~\cite{Roberts2020HypersimAP}, we compare SAM-V with SAM2~\cite{Ravi2024SAM2S} in a controlled prompt-conditioned setting. 
This evaluation directly targets \textbf{Q1}: segmenting the prompted object across views where it may disappear and reappear at different image locations due to camera motion or occlusion.

For each evaluated instance, we construct an input sequence of $8$ positive frames, where the target is visible, and $8$ negative frames, where it is absent, randomly interleaved. Positive frames are selected under two settings: \emph{continuous}, using nearest-point sampling over camera poses for small viewpoint changes, and \emph{diverse}, using farthest-point sampling for large viewpoint variation. The two settings disentangle the sources of difficulty: continuous sampling tests temporal understanding in a video-like regime, while diverse sampling isolates large viewpoint changes, where appearance-based association is expected to fail. Both methods receive the same prompts: $10$ points sampled on the target in one randomly chosen positive frame. For SAM2, the prompted frame serves as the anchor, and masks are propagated forward and backward through the sequence.

As shown in Table~\ref{tab:hypersim_overall_iou}, our method achieves a large improvement over SAM2 under the diverse setting, improving the O-IoU from $48.1$ to $62.4$, and the R@50 from $59.6$ to $71.9$. This suggests that incorporating VGGT geometry features is particularly beneficial when the input views exhibit large camera pose variation. Under the continuous setting, both methods perform similarly.

In Fig.~\ref{fig:qual_sam2_vggtsam}, we show that SAM2 struggles to maintain the target identity in challenging multi-view scenes. In the scene with multiple visually similar objects, SAM2 often confuses the prompted target with nearby instances. In the examples where the target undergoes significant appearance changes and may temporarily disappear before reappearing, SAM2 can lose track of the object. In contrast, SAM-V leverages multi-view geometric cues from VGGT together with SAM's segmentation prior, enabling it to produce accurate and consistent masks across views under both instance ambiguity and large camera motion.

\begin{table*}[htbp]
\vspace{-.5em}
\centering
\caption{\textbf{Comparison on ScanNet++ and ScanNet 3D tracking benchmarks.}
\textsuperscript{*}~denotes IGGT results from our own evaluation protocol, which differs from the originally reported numbers and was verified through correspondence with the authors. Time is the average end-to-end inference time per scene (6--9 images, one NVIDIA L40S), from RGB inputs to final object masks; for SAM2, Point-SAM, and SAM-V this includes proposal generation, and for ODIN this includes point regression from VGGT. Bold marks the best result. }
\label{tab:iggt_tracking_results}
\small
\resizebox{1.0\linewidth}{!}{
\begin{tabular}{lcccccccccc}
\toprule
& \multicolumn{5}{c}{\textbf{ScanNet++}} 
& \multicolumn{5}{c}{\textbf{ScanNet (Zero-Shot)}} \\
\cmidrule(lr){2-6} \cmidrule(lr){7-11}
Method 
& T-mIoU & O-IoU & P@50 & R@50 & Time (s)
& T-mIoU & O-IoU & P@50 & R@50 & Time (s) \\
\midrule
SAM2 
& 65.9 & 67.8 & 77.7 & 68.9 & 40.1
& 69.4 & 70.0 & 72.7 & 72.3 & 30.2 \\
IGGT 
& 65.5\textsuperscript{*} & 67.9 & 59.6 & 71.7 & 223.1
& 63.3\textsuperscript{*} & 63.6 & 52.7 & 73.2 & 252.4 \\
PanSt3R 
& 72.7 & 74.8 & 77.4 & 77.4 & 4.8
& 71.2 & 73.3 & 72.1 & 80.8 & 4.9 \\
ODIN 
& 52.1 & 55.5 & 51.7 & 57.1 & 4.6
& 58.1 & 63.1 & 63.7 & 69.6 & 4.4 \\
Point-SAM 
& 41.4 & 45.0 & 40.2 & 42.5 & 229.4
& 50.6 & 57.4 & 62.5 & 60.3 & 190.5 \\
\midrule
SAM-V (Ours)
& \textbf{78.4} & \textbf{79.4} & \textbf{79.4} & \textbf{89.2} & 49.4
& \textbf{75.6} & \textbf{77.3} & \textbf{77.4} & \textbf{87.1} & 38.6 \\
\bottomrule
\end{tabular}
}
\end{table*}

\begin{figure*}[!htbp]
    \centering
    \includegraphics[width=\linewidth]{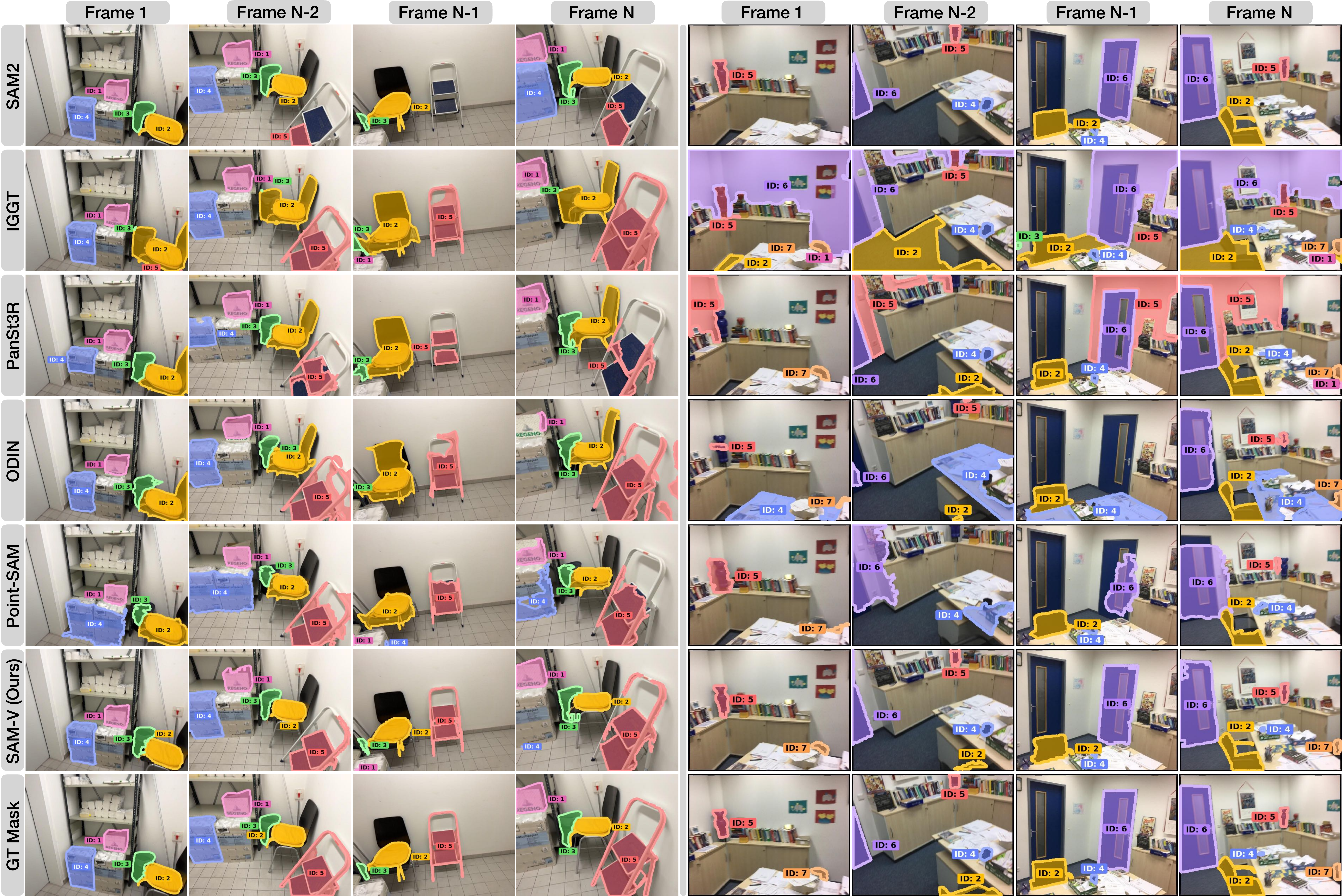}
    \caption{
    \textbf{Qualitative comparison on the IGGT 3D tracking benchmark.} 
    Two scenes of nine frames each; four frames per scene are shown for space. Colors and IDs denote cross-view identity.
    SAM-V produces more accurate and consistent multi-view object masks, while the baselines can fail under challenging cases such as target ambiguity, large viewpoint changes, or incomplete cross-view tracking.
    }
    \label{fig:qual_compare_iggt}
\end{figure*}

\paragraph{Every object segmentation.}
We further evaluate SAM-V on the ScanNet++~\cite{Yeshwanth2023ScanNetAH} and ScanNet~\cite{Dai2017ScanNetR3} scenes from the IGGT 3D tracking benchmark~\cite{Li2025IGGTIG}.
The predicted mask tracks are matched to ground-truth object tracks using Hungarian matching based on O-IoU. 
This setting targets \textbf{Q2}: unlike the prompt-conditioned Hypersim evaluation, the benchmark requires segmenting every object without user-specified prompts, so we evaluate SAM-V with the every-object inference procedure of Sec.~\ref{sec:every_object_inference}.

Table~\ref{tab:iggt_tracking_results} compares SAM-V with SAM2, IGGT, PanSt3R, ODIN, and Point-SAM. On ScanNet++, SAM-V outperforms IGGT by $12.9$ T-mIoU, $11.5$ O-IoU, $19.8$ P@50, and $17.5$ R@50, and surpasses PanSt3R, the strongest baseline, by $5.7$ T-mIoU and $11.8$ R@50. On the zero-shot ScanNet evaluation, SAM-V again leads on all four metrics, indicating that the learned geometry-aware fusion transfers to unseen real-world scenes beyond the training domain. Query-based methods (PanSt3R, ODIN) are efficient, since instances are decoded
from a single shared set of learnable queries over all input views, but this comes
at a cost in recall: scene-level queries discover instances at a granularity that
does not always match the annotation, most often under-segmenting. A representative
case is predicting an entire chair where the target is an item placed on it
(Fig.~\ref{fig:qual_compare_iggt}). SAM-V instead derives prompts from SAM
proposals and thus inherits SAM's instance granularity. Promptability also matters beyond the benchmark: PanSt3R and ODIN discover instances
autonomously, so the target instance cannot be specified at inference time, whereas SAM-V supports both prompted and every-object modes in one model.

Fig.~\ref{fig:qual_compare_iggt} shows that each baseline exhibits a distinct
failure mode. SAM2 often struggles to maintain the target identity across views because it primarily relies on image-level appearance and temporal propagation. IGGT may still produce incomplete or inconsistent masks when the object undergoes large viewpoint changes or appears in cluttered scenes. PanSt3R and ODIN additionally show granularity mismatches,  merging small objects into the larger structures: the bag
(ID: 2) into the chair in the left scene, and the cup (ID: 4) into the table in
the right scene. SAM-V avoids all three failure modes and maintains object identity across views.

\begin{figure*}[htbp]
    \centering
    \includegraphics[width=\linewidth]{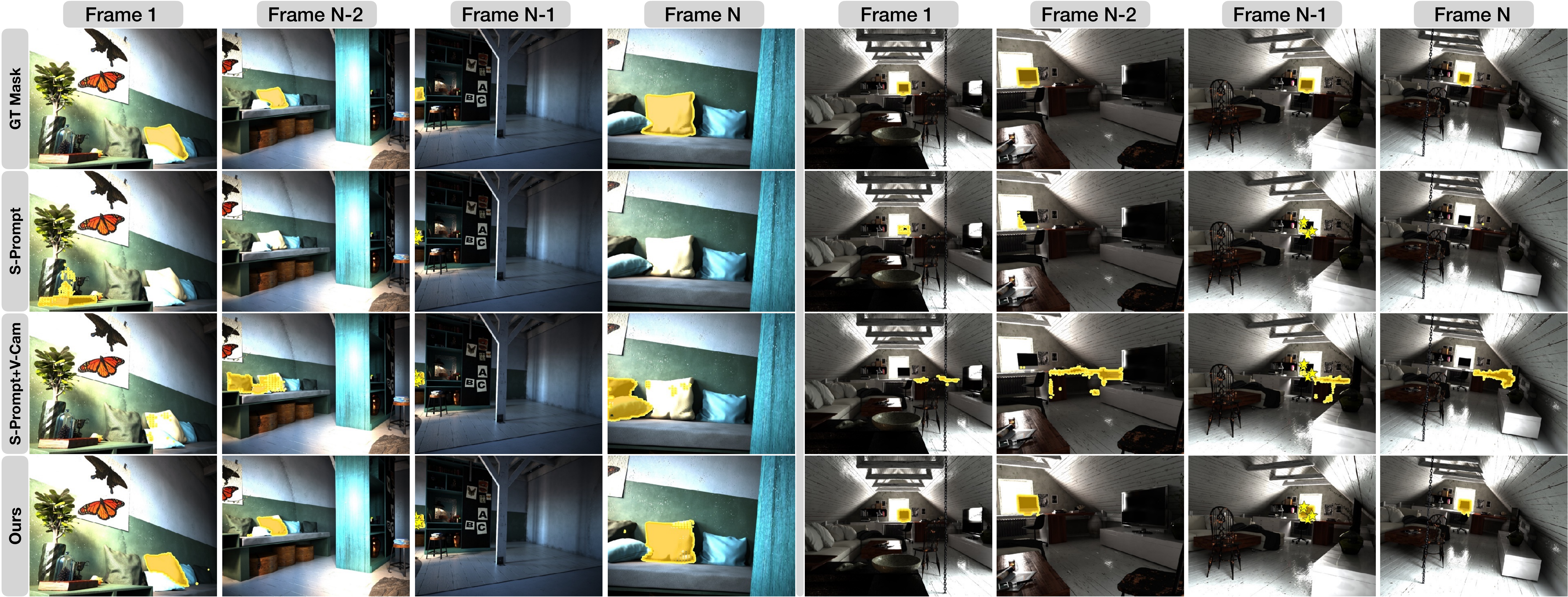}
\caption{
\textbf{Qualitative ablation of prompt-fusion inputs.}
We compare SAM-Prompt, SAM-Prompt+VGGT-Cam, and SAM-Prompt+VGGT-Cam+VGGT-Feat on two Hypersim scenes.
Using only SAM prompt tokens or adding only the VGGT camera token can lead to prediction shifts toward nearby regions, while further incorporating the VGGT local feature sampled improves spatial grounding and produces more accurate masks across views.
}
    \label{fig:ablate_patch}
\end{figure*}

\subsection{Ablation Study}
\label{sec:ablation}

We ablate the prompt-fusion module by comparing three variants with progressively richer prompt representations:
(1) \emph{SAM-Prompt}, using only the original SAM prompt embeddings;
(2) \emph{SAM-Prompt + Cam.}, which fuses the SAM prompt embeddings with the VGGT camera token of the prompted view; and
(3) \emph{SAM-Prompt + Cam. + Feat.}, our full design, which further incorporates the local VGGT feature at the prompted coordinate.
This ablation answers \textbf{Q3}: how much the view-level and point-level geometric context each contribute to grounding sparse prompts.

\begin{table}[t]
\centering
\caption{
\textbf{Ablation study of prompt-fusion inputs.}
We evaluate progressively richer prompt representations under continuous and diverse view sampling on the Hypersim test split.
\textit{Cam.} and \textit{Feat.} refer to camera tokens and local feature tokens from VGGT.
}
\label{tab:ablation_patch}
\small
\resizebox{\linewidth}{!}{
\begin{tabular}{lcccccc}
\toprule
& \multicolumn{3}{c}{\textbf{Continuous}} 
& \multicolumn{3}{c}{\textbf{Diverse}} \\
\cmidrule(lr){2-4} \cmidrule(lr){5-7}
Method 
& O-IoU & P@50 & R@50
& O-IoU & P@50 & R@50 \\
\midrule
SAM-Prompt
& 21.3 & 16.3 & 21.9
& 23.0 & 17.4 & 23.5 \\
SAM-Prompt + Cam. 
& 35.5 & 29.9 & 34.5
& 32.2 & 25.9 & 29.9 \\
SAM-Prompt + Cam. + Feat.
& \textbf{64.2} & \textbf{69.3} & \textbf{75.5}
& \textbf{62.4} & \textbf{65.1} & \textbf{71.9} \\
\bottomrule
\end{tabular}
}
\end{table}

As shown in Table~\ref{tab:ablation_patch}, each added component brings a large gain under both continuous and diverse sampling, with the full design improving O-IoU from $21.3$ to $64.2$ (continuous) and from $23.0$ to $62.4$ (diverse).
Fig.~\ref{fig:ablate_patch} illustrates the mechanism behind these gains.
With SAM prompt embeddings alone, the model cannot tell which view the prompt refers to and fails to locate the correct object.
Adding the camera token resolves the view: the model consistently segments an object near the queried one across all views, but the camera token alone does not pinpoint the exact location, so predictions shift to nearby instances.
Additionally sampling the local VGGT feature at the prompted coordinate links the prompt embedding directly to the dense multi-view image representation, yielding accurate and consistent masks across views.

\section{Conclusion}
\vspace{+.7em}

We introduce SAM-V, an end-to-end framework for multi-view consistent instance segmentation that bridges 2D segmentation priors with 3D geometric understanding. By integrating SAM and VGGT representations through geometry-aware image and prompt fusion, SAM-V predicts accurate masks across views in a single forward pass, and extends from prompt-conditioned segmentation to every-object scene-level inference within one model. Evaluations on Hypersim, ScanNet++, and ScanNet show SAM-V consistently outperforms baselines, maintaining object identity under severe viewpoint changes. Furthermore, ablations confirm that explicitly anchoring sparse prompts with local VGGT features is critical for resolving spatial ambiguity.

\vspace{+.3em}
\paragraph{Limitations.}
Currently SAM-V is trained only on indoor datasets (Hypersim and ScanNet++), and generalization to outdoor or object-centric
domains remains untested. Our formulation assumes rigid objects in static environments; handling deformable entities or dynamic scenes will require spatial-temporal modeling for non-rigid motion. Finally, SAM-V focuses on prompt-conditioned tracking rather than open-vocabulary or language-guided segmentation. Overcoming these limitations will further optimize these lightweight 3D inference algorithms for downstream robotic applications, such as real-time humanoid control and household organizing tasks.

\paragraph{Acknowledgement.}
This work is supported in part by NSF IIS grant 2312102.
S.W. is supported by NSF 2331878 and 2340254, and research grants from Intel, Amazon, and IBM. 
This research used both the DeltaAI advanced computing and data resource, which is supported by the National Science Foundation (award OAC 2320345) and the State of Illinois, and the Delta advanced computing and data resource which is supported by the National Science Foundation (award OAC 2005572) and the State of Illinois. Delta and DeltaAI are joint efforts of the University of Illinois Urbana-Champaign and its National Center for Supercomputing Applications.

\clearpage

{
    \small
    \bibliographystyle{plain}
    \bibliography{main}
}

\clearpage
\appendix

\setcounter{figure}{0}
\setcounter{table}{0}
\renewcommand{\thefigure}{A\arabic{figure}}
\renewcommand{\thetable}{A\arabic{table}}

\section{Additional Implementation and Training Details}
\label{sec:supp_training_details}

\subsection{Model and Optimization}

SAM-V uses the ViT-H variant of SAM as the 2D segmentation backbone. The SAM image encoder, the SAM prompt encoder, and the VGGT encoder are frozen throughout training; the trainable components are the image feature fusion MLP, the prompt fusion cross-attention module, and the SAM mask decoder, which is initialized from the pretrained SAM weights. This design preserves the pretrained 2D segmentation prior from SAM and the multi-view geometric representation from VGGT, while adapting only the fusion modules and the decoder for multi-view consistent segmentation.

All input frames are resized to $1024 \times 1024$. Training uses groups of $N=8$ frames, four positive and four negative. The prompt-conditioned comparison with SAM2 on Hypersim uses $N=16$ frames per object, eight positive and eight negative; the every-object evaluation uses all frames of a scene, which is six to nine on the benchmark splits. The output mask is predicted over the horizontally concatenated multi-view image layout, giving a mask of spatial size $H \times (NW)$.

Optimization settings are summarized in Tab.~\ref{tab:training_hyperparams}. Two details are not captured there. For the focal loss, we follow the point-sampling strategy of SAM-HQ and sample $112 \times 112$ supervision points per frame, so the loss is evaluated on $8 \times 112 \times 112$ locations per training sample, while the Dice loss is computed over the full concatenated multi-view mask. The additional term $\mathcal{L}_{\mathrm{IoU}}$ used in the second stage is the mean squared error between the decoder's predicted mask quality and the achieved IoU, following the original SAM objective; it supervises the score by which candidate masks are ranked during every-object inference.

\begin{table}[htbp]
\centering
\caption{\textbf{Training configuration of SAM-V.}
Settings shared by both stages are shown once, while stage-specific
sampling and loss configurations are reported separately.}
\label{tab:training_hyperparams}

\footnotesize
\setlength{\tabcolsep}{3pt}
\renewcommand{\arraystretch}{1.05}

\begin{tabular}{@{}
>{\raggedright\arraybackslash}p{0.25\columnwidth}
>{\raggedright\arraybackslash}p{0.345\columnwidth}
>{\raggedright\arraybackslash}p{0.345\columnwidth}
@{}}
\toprule
Hyperparameter
& Hypersim training
& ScanNet++ finetuning
\tabularnewline
\midrule

SAM backbone
& \multicolumn{2}{>{\raggedright\arraybackslash}p{0.69\columnwidth}}{ViT-H}
\tabularnewline

Input resolution
& \multicolumn{2}{>{\raggedright\arraybackslash}p{0.69\columnwidth}}{$1024 \times 1024$}
\tabularnewline

Frames per sample
& \multicolumn{2}{>{\raggedright\arraybackslash}p{0.69\columnwidth}}{$8$ ($4$ positive and $4$ negative)}
\tabularnewline

Optimizer
& \multicolumn{2}{>{\raggedright\arraybackslash}p{0.69\columnwidth}}{AdamW}
\tabularnewline

Learning rate
& \multicolumn{2}{>{\raggedright\arraybackslash}p{0.69\columnwidth}}{$1 \times 10^{-4}$}
\tabularnewline

Weight decay
& \multicolumn{2}{>{\raggedright\arraybackslash}p{0.69\columnwidth}}{$0.01$}
\tabularnewline

Batch size
& \multicolumn{2}{>{\raggedright\arraybackslash}p{0.69\columnwidth}}{$2$}
\tabularnewline

Scheduler
& \multicolumn{2}{>{\raggedright\arraybackslash}p{0.69\columnwidth}}{Cosine annealing, $T_{\max}=200$, $\eta_{\min}=1 \times 10^{-6}$}
\tabularnewline

Prompt points
& $20$
& Randomly sampled from $\{3,4,5\}$
\tabularnewline

Prompted frames
& Progressive multi- to single-frame
& Single frame
\tabularnewline

Positive sampling
& Progressive nearest to farthest
& Farthest-point sampling
\tabularnewline

Loss
& $20\mathcal{L}_{\mathrm{focal}} + \mathcal{L}_{\mathrm{dice}}$
& $20\mathcal{L}_{\mathrm{focal}}+\mathcal{L}_{\mathrm{dice}}+ \mathcal{L}_{\mathrm{IoU}}$
\tabularnewline

Training scenes
& $614$
& $856$
\tabularnewline

Training epochs
& $212$ (checkpoint at $206$)
& $20$
\tabularnewline

Hardware
& $8$ NVIDIA L40S GPUs
& $14$ NVIDIA L40S GPUs
\tabularnewline

\bottomrule
\end{tabular}
\end{table}

\subsection{Two-Stage Training Schedule}

SAM-V is trained on Hypersim for $212$ epochs, and the checkpoint at epoch $206$, which achieves the highest validation IoU, initializes the second stage. ScanNet++ finetuning then runs for $20$ epochs, again selecting the checkpoint by validation IoU.

Each training sample consists of eight frames associated with one target object. A positive frame is one in which the target occupies at least $0.0025$ of the image area; a negative frame is one in which the target instance is absent; frames in which the target is present but falls below the area threshold are excluded from both sets. Four positive frames and four randomly sampled negative frames form the sample, so the model is explicitly supervised to predict all-zero masks in views without the target.

The Hypersim stage applies a progressive sampling curriculum over its first $70$ epochs, during which two probabilities increase linearly from $0$ to $1$. The first is the probability of selecting positive frames by farthest-point sampling over their camera poses rather than nearest-point sampling, which gradually widens the viewpoint spread within a sample. The second is the probability of drawing all prompt points from a single positive frame rather than distributing them across the four. After epoch $70$ every sample uses the final configuration: four pose-diverse positive frames, four negative frames, and all prompt points on one positive frame. The curriculum begins with nearby views and prompt evidence spread over several frames, and shifts toward the harder setting in which object identity must propagate from a single prompted view to geometrically diverse ones while predictions are suppressed in negative views.

ScanNet++ finetuning starts from this final configuration and does not repeat the curriculum. It differs from the first stage in two respects: the number of prompt points per sample is drawn from $\{3,4,5\}$ rather than fixed at $20$, exposing the model to varying amounts of sparse prompt evidence, and the objective adds $\mathcal{L}_{\mathrm{IoU}}$, with the focal, Dice, and IoU-prediction terms weighted $20{:}1{:}1$.

Prompt counts differ between training and inference. The prompt-conditioned Hypersim evaluation uses $10$ points on a randomly selected positive frame, given identically to SAM-V and SAM2, and every-object inference samples $5$ points inside each SAM proposal.

\section{Every-Object Inference Details}
\label{sec:supp_everyobject}

SAM-V is prompt-conditioned by construction: one prompt group produces one multi-view mask. Every-object inference turns it into a scene-level segmenter by generating prompt groups automatically and deduplicating the resulting multi-view masks. The procedure has three stages: proposal generation, prompt sampling, and multi-view filtering.

\subsection{Proposal Generation}

For each of the $N$ input frames we run the SAM automatic mask generator independently to obtain 2D proposal masks. The generator uses the ViT-H backbone with a default configuration of $32 \times 32$ point grid, $64$ points per batch, a predicted-IoU threshold of $0.88$, a stability-score threshold of $0.95$, a box-NMS threshold of $0.7$, and no minimum mask region area. All multimask output levels are retained. Proposals are then sorted by predicted IoU, stability score, and area in descending order, and at most $64$ proposals per frame are kept. This stage is shared with the Point-SAM baseline, which consumes the same proposals through the same sampler (Sec.~\ref{sec:supp_baselines}).

\subsection{Prompt Sampling}

Each retained proposal is converted into one prompt group of $5$ all-positive points sampled inside its mask region. Points are selected by a pole-plus-diverse strategy: the first point is placed at the pole of inaccessibility of the mask, that is, the pixel maximizing the distance transform to the mask boundary; each subsequent point greedily maximizes
\[
d_{\min}^{2}(x) + 0.5 \cdot \mathrm{dt}^{2}(x),
\]
where $d_{\min}(x)$ is the distance from $x$ to the already selected points and $\mathrm{dt}(x)$ is the distance transform value at $x$. Candidates are restricted to pixels with $\mathrm{dt}(x) \geq \max(1,\, 0.35\,\mathrm{dt}_{\max})$, which keeps prompts away from mask boundaries. This favors points that are interior to the proposal and spread across it, which stabilizes the decoded mask for elongated and concave objects.

\subsection{Multi-View Filtering and Deduplication}

Every prompt group is decoded into one multi-view mask spanning all $N$ frames. Candidates then pass through a filter chain: the decoder's predicted IoU is thresholded at $0.40$; the stability score, computed at mask threshold $0.0$ with offset $1.0$, is thresholded at $0.20$; and the remaining logits are binarized. Surviving candidates are converted to multi-view masks and deduplicated by non-maximum suppression with an overlap threshold of $0.95$, ranked by the decoder's predicted IoU. Because each candidate already spans all views, duplicates of the same physical object coincide across the entire multi-view volume, so overlap is measured on the concatenated multi-view mask rather than per frame. The surviving masks form the final every-object output, and their enumeration order defines the object identities used for evaluation.

Inference uses $8$ prompt groups per mask-decoder call, and all frames are resized to $1024 \times 1024$, matching the training resolution.

\section{Evaluation Protocol and Metric Definitions}
\label{sec:supp_metrics}

This section gives the formal definitions of the metrics reported in the main paper. All metrics are computed per scene and then aggregated over the scenes of a split.

\subsection{Notation}

A scene has $N$ frames. Let $\mathcal{G} = \{M_1, \dots, M_{|\mathcal{G}|}\}$ be the ground-truth object tracks, where $M_g = (M_g^1, \dots, M_g^N)$ and $M_g^i \in \{0,1\}^{H \times W}$ is the mask of object $g$ in frame $i$, empty when the object is not visible. Let $\mathcal{P} = \{\hat{M}_1, \dots, \hat{M}_{|\mathcal{P}|}\}$ be the predicted tracks in the same form. Ground-truth object identities are taken as the positive labels present in the ground-truth maps; both images and label maps are resized to $1024 \times 1024$, with nearest-neighbor interpolation for labels.

\subsection{Per-Frame and Multi-View IoU}

The per-frame IoU between a predicted track $\hat{M}_p$ and a ground-truth track $M_g$ at frame $i$ is
\[
\mathrm{IoU}^i(\hat{M}_p, M_g) =
\frac{|\hat{M}_p^i \cap M_g^i|}{|\hat{M}_p^i \cup M_g^i|},
\]
defined as $1$ when both masks are empty. The multi-view IoU, reported as O-IoU, is a single IoU over the stacked multi-view volume,
\[
\mathrm{O\text{-}IoU}(\hat{M}_p, M_g) =
\frac{\sum_{i=1}^{N} |\hat{M}_p^i \cap M_g^i|}
     {\sum_{i=1}^{N} |\hat{M}_p^i \cup M_g^i|},
\]
which differs from the mean of per-frame IoUs in that a frame contributes in proportion to its mask area, and a prediction that segments the correct object in one frame and a different object in another is penalized by the union in both frames.

\subsection{Track Matching}

Predicted tracks are assigned to ground-truth tracks by one-to-one Hungarian matching on the pairwise O-IoU matrix, maximizing the total O-IoU. Pairs with zero O-IoU are discarded after the assignment, so a ground-truth track receives a match only if some predicted track overlaps it in at least one frame. Ground-truth tracks left without a match are retained in the evaluation and scored as zero for T-mIoU and O-IoU.

\subsection{Frame-Level Precision and Recall}

Precision and recall are counted at the level of individual frames, at IoU thresholds $\tau \in \{0.1, \dots, 0.9\}$. For each matched pair $(\hat{M}_p, M_g)$ and each frame $i \in \{1, \dots, N\}$:

\begin{itemize}
\itemsep0em
\item if $M_g^i$ and $\hat{M}_p^i$ are both non-empty, the frame is a true positive when $\mathrm{IoU}^i \geq \tau$, and a false positive together with a false negative otherwise;
\item if $M_g^i$ is non-empty and $\hat{M}_p^i$ is empty, the frame is a false negative;
\item if $\hat{M}_p^i$ is non-empty and $M_g^i$ is empty, the frame is a false positive.
\end{itemize}

For an unmatched ground-truth track, every visible frame is a false negative. Predicted tracks that the assignment leaves unmatched do not contribute counts. Precision and recall are then
\[
P@\tau = \frac{\mathrm{TP}}{\mathrm{TP} + \mathrm{FP}},
\qquad
R@\tau = \frac{\mathrm{TP}}{\mathrm{TP} + \mathrm{FN}},
\]
and are $0$ when the denominator is $0$. P@50 and R@50 denote $\tau = 0.5$.

The third case above is what penalizes a prediction that leaks into frames where the target is absent, which is the failure mode the negative frames in training are meant to suppress. The second case penalizes a track that drops the object in a view where it remains visible.

\subsection{Aggregation Over a Split}

T-mIoU and O-IoU are micro-averaged over all ground-truth objects of all scenes in the split, so every object carries equal weight regardless of scene size. Precision and recall are computed once from the true-positive, false-positive, and false-negative counts pooled over all scenes, rather than as an average of per-scene ratios.

\section{Baseline Configurations}
\label{sec:supp_baselines}

Each baseline produces multi-view object tracks in a common format, which are then scored by the protocol of Sec.~\ref{sec:supp_metrics}. All methods receive RGB images alone; none is given sensor depth or ground-truth camera poses. ODIN and Point-SAM operate on 3D inputs, so both are supplied with pointmaps predicted by VGGT-1B from the same images.

\subsection{SAM2}

We use the off-the-shelf SAM 2.1 Hiera-Large checkpoint. Frames are resized to $1024 \times 1024$ and treated as a video sequence. In the prompt-conditioned Hypersim evaluation, the prompted frame is used as the anchor and the mask is propagated both forward and backward through the sequence, with the multi-mask branch disabled and occlusion prediction enabled. For the every-object evaluation on the 3D tracking benchmark, initial masks are obtained from SAM2's automatic mask generation on the first frame and propagated through the remaining frames, and each propagated mask is one predicted track.

\subsection{IGGT}

IGGT is evaluated with the official released model. The numbers we report come from our own evaluation protocol rather than the originally published figures, so that IGGT is scored by the same matching and metric definitions as every other method. Object tracks are formed by the model's own instance grouping without a user-specified prompt.

\subsection{PanSt3R}

We use the released \texttt{panst3r\_v2\_512\_5ds} checkpoint with an input size of $512$, the standard v2 post-processing, and bfloat16 inference; the optional cuRoPE extension is not installed. The class vocabulary is upstream's own demo default, a $100$-name list corresponding to the ScanNet++ semantic benchmark, used unchanged on both splits. Predicted segments are grouped into cross-view tracks by query identity. Because PanSt3R computes its panoptic assignment jointly over all input views, its tracks are multi-view consistent by construction, which is what allows them to enter the track-matching stage directly. PanSt3R discovers objects through learnable scene-level queries and has no prompt-conditioned mode, so it appears only in the every-object comparison.

\subsection{ODIN}

We use the ScanNet200 instance-segmentation checkpoint with the Swin-B backbone. ODIN requires posed RGB-D input, which the benchmark does not provide, so it is given VGGT pointmaps that are gravity-aligned and rescaled per scene by $s = 1.55\,\mathrm{m} / h_{\mathrm{cam}}$, where $h_{\mathrm{cam}}$ is the camera height estimated from the predicted geometry. A per-scene convention is required because the true scale varies by a factor of $2.63$ across the benchmark scenes, so any single global constant would be substantially wrong on some of them. Ghost points and precomputed segments are disabled, since both require the ScanNet mesh; the checkpoint was trained with them enabled, and this configuration difference is a known limitation of the comparison. Predicted tracks are ODIN's $100$ object queries, one track per query taken by maximum over classes, with no additional suppression, as ODIN is a set predictor.

We report the checkpoint trained on the $200$-class ScanNet200 vocabulary rather than the $20$-class variant, since the narrower vocabulary suppresses recall on objects outside its label set in a class-agnostic benchmark. Even so, ScanNet200 names $77.6\%$ of the ground-truth objects on ScanNet++ and $89.4\%$ on ScanNet. The matching stage never reads a predicted class label, so objects outside the vocabulary can still be matched and scored.

\subsection{Point-SAM}

We use the ViT-L Point-SAM model. Its geometry comes from VGGT-predicted pointmaps sampled on a stride-$4$ pixel grid and voxel-merged within each prompt's crop; predicted 3D masks are returned to 2D by a voxel-to-pixel lookup rather than by rendering. Prompts are the same SAM proposals and the same pole-plus-diverse sampler with $5$ points used by SAM-V's every-object inference (Sec.~\ref{sec:supp_everyobject}), so the two methods differ in how a prompt is resolved rather than in where prompts come from. Both mask selection among Point-SAM's output heads and the ranking used for suppression are by stability score.

\section{Timing Measurement Protocol}
\label{sec:supp_timing}

The runtimes in Tab.~\ref{tab:iggt_tracking_results} of the main paper are strictly serial, single-GPU measurements on one NVIDIA L40S with no other work on the node, reported as the mean over the scenes of a split. Each figure covers the full path from RGB inputs to final object masks. Timings measured on other hardware are not comparable to these.

Table~\ref{tab:supp_timing} breaks the per-scene time into stages. SAM-V and Point-SAM share the same SAM proposal generation (Sec.~\ref{sec:supp_everyobject}), which is charged to both.

\begin{table}[htbp]
\centering
\caption{\textbf{Per-scene inference time by stage.}
All measurements are serial on one NVIDIA L40S, averaged over the scenes of
each split. Scenes contain $6$--$9$ frames. Prep.\ denotes proposal generation
for SAM-V and Point-SAM and the VGGT forward pass for ODIN; Asm.\ denotes
resizing and mask encoding.}
\label{tab:supp_timing}
\footnotesize
\setlength{\tabcolsep}{4pt}
\begin{tabular}{@{}llrrrr@{}}
\toprule
Method & Split & Prep. & Infer & Asm. & Total \tabularnewline
\midrule
SAM-V & ScanNet++ & $33.1$ & $16.3$ & -- & $49.4$ \tabularnewline
SAM-V & ScanNet & $22.2$ & $16.3$ & -- & $38.6$ \tabularnewline
PanSt3R & ScanNet++ & -- & $2.6$ & $2.2$ & $4.8$ \tabularnewline
PanSt3R & ScanNet & -- & $2.8$ & $2.1$ & $4.9$ \tabularnewline
ODIN & ScanNet++ & $2.4$ & $0.5$ & $1.8$ & $4.6$ \tabularnewline
ODIN & ScanNet & $2.2$ & $0.5$ & $1.7$ & $4.4$ \tabularnewline
Point-SAM & ScanNet++ & $37.3$ & $176.6$ & $9.4$ & $229.4$ \tabularnewline
Point-SAM & ScanNet & $25.6$ & $157.5$ & $6.6$ & $190.5$ \tabularnewline
\bottomrule
\end{tabular}
\end{table}

Two observations follow from the breakdown. First, SAM-V's own multi-view decoding takes $16.3$\,s on both splits, and the shared proposal stage accounts for $57$--$67\%$ of its total; PanSt3R and ODIN predict their own masks and pay none of that cost. Second, only $0.5$\,s of ODIN's runtime is ODIN itself, with roughly half of the remainder spent on the VGGT geometry it cannot produce on its own.

Timings for SAM-V, Point-SAM, and ODIN use \texttt{perf\_counter} with CUDA synchronization and two discarded warmup scenes. The PanSt3R adapter transfers its output to CPU, which forces a synchronization, so its total is measured reliably, but its internal stage split is not resolved at the same granularity. The difference between the first scene and the remaining scenes is negligible on both splits.

\section{Additional Quantitative Results}
\label{sec:supp_additional_quantitative}

\begin{figure*}[!htb]
    \centering

    \begin{subfigure}[t]{0.48\linewidth}
        \centering
        \includegraphics[width=\linewidth]{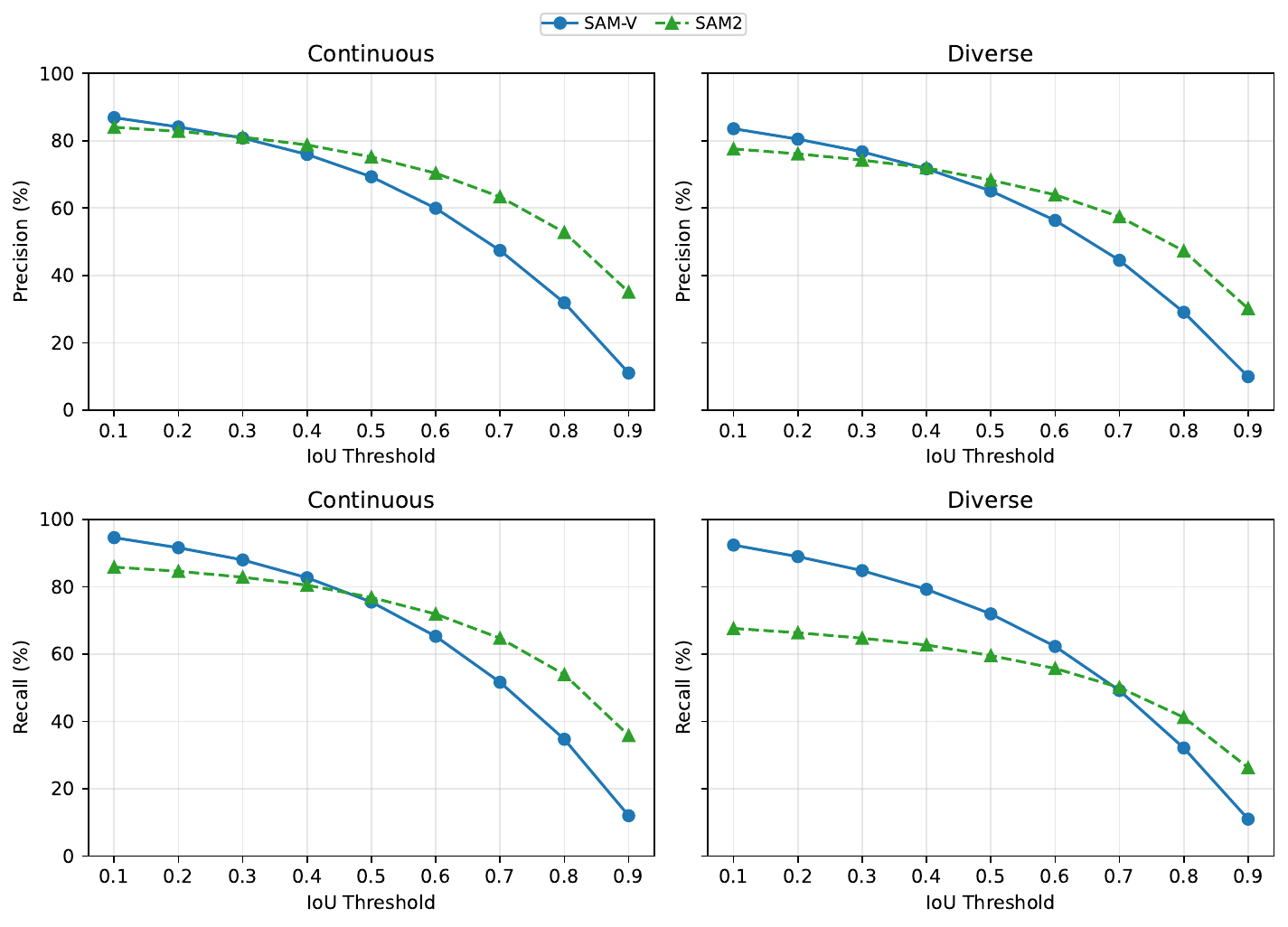}
        \caption{Comparison with SAM2 on Hypersim. Continuous and Diverse are two different image sampling methods.}
        \label{fig:hypersim_pr}
    \end{subfigure}
    \hfill
    \begin{subfigure}[t]{0.48\linewidth}
        \centering
        \includegraphics[width=\linewidth]{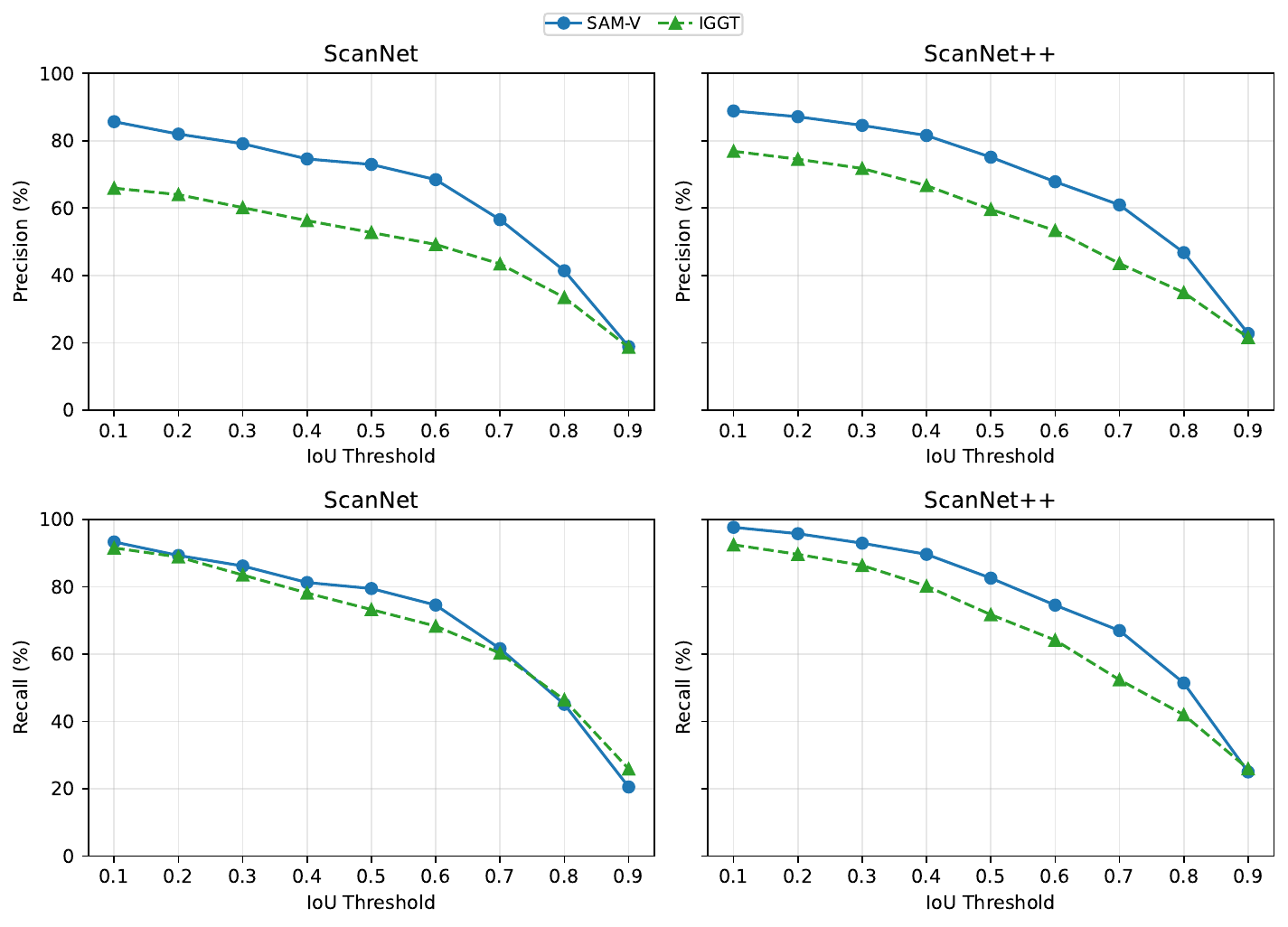}
        \caption{Comparison with IGGT on ScanNet and ScanNet++.}
        \label{fig:scannet_pr}
    \end{subfigure}

    \caption{Precision-recall comparisons across datasets and evaluation protocols.}
    \label{fig:precision_recall_comparison}
\end{figure*}

In addition to the main quantitative comparisons reported in the paper, we provide precision-recall curves under different IoU thresholds in Fig.~\ref{fig:precision_recall_comparison}. These curves offer a more complete view of model performance beyond a single IoU threshold, showing how precision and recall change as the matching criterion becomes stricter.

On Hypersim, we compare SAM-V with SAM2 under both continuous and diverse frame sampling protocols. The continuous setting evaluates nearby frames sampled from the same scene, while the diverse setting uses frames with larger viewpoint variation. This comparison tests whether a method can maintain object consistency not only under temporally or spatially adjacent views, but also under more challenging cross-view changes.

On ScanNet and ScanNet++, we compare SAM-V with IGGT using the same tracking-style evaluation protocol as in the main paper. The precision-recall curves further show the trade-off between accurate object discovery and false positive suppression across IoU thresholds. Overall, these additional results complement the main benchmark tables and provide a finer-grained analysis of multi-view consistent segmentation quality.

\section{Objects Outside the Classification Vocabulary}
\label{sec:supp_ood}

The every-object comparison in the main paper is run on benchmark scenes whose
objects are largely drawn from standard indoor categories. This section examines
a scene assembled to fall outside that regime, in order to characterize how the
two families of methods behave when the objects present are not nameable in a
fixed class list.

\subsection{Scene and Annotation}

We captured eight handheld views of a tabletop arrangement holding sixteen
objects, including a ukulele, three plush toys, a camera, a projector, a pair of
goggles, two game controllers, a power strip, and a knife. Instance masks were
annotated interactively for every object in every view in which it appears, at
$1024 \times 1024$, with each frame confirmed manually; object identities are
consistent across the eight views by construction. Category names are recorded
for reference only. Neither method reads them: SAM-V is prompted from points and
PanSt3R is conditioned on its own runtime class list.

\subsection{Setup}

PanSt3R is evaluated under three class vocabularies and their union, with the
released checkpoint and post-processing held fixed in all four runs. PanSt3R-S
uses the upstream demo default, a $100$-name list corresponding to the ScanNet++
semantic benchmark, which is the configuration used for Tab.~\ref{tab:iggt_tracking_results} of the main
paper. PanSt3R-C uses the $133$-class COCO panoptic list and PanSt3R-A the
$150$-class ADE20K list. The union of the three, $316$ names, corresponds to the
vocabulary over which the released checkpoint was trained. SAM-V is run in the
prompt-conditioned mode on the same eight frames.

\begin{figure*}[htbp]
\centering
\includegraphics[width=\linewidth]{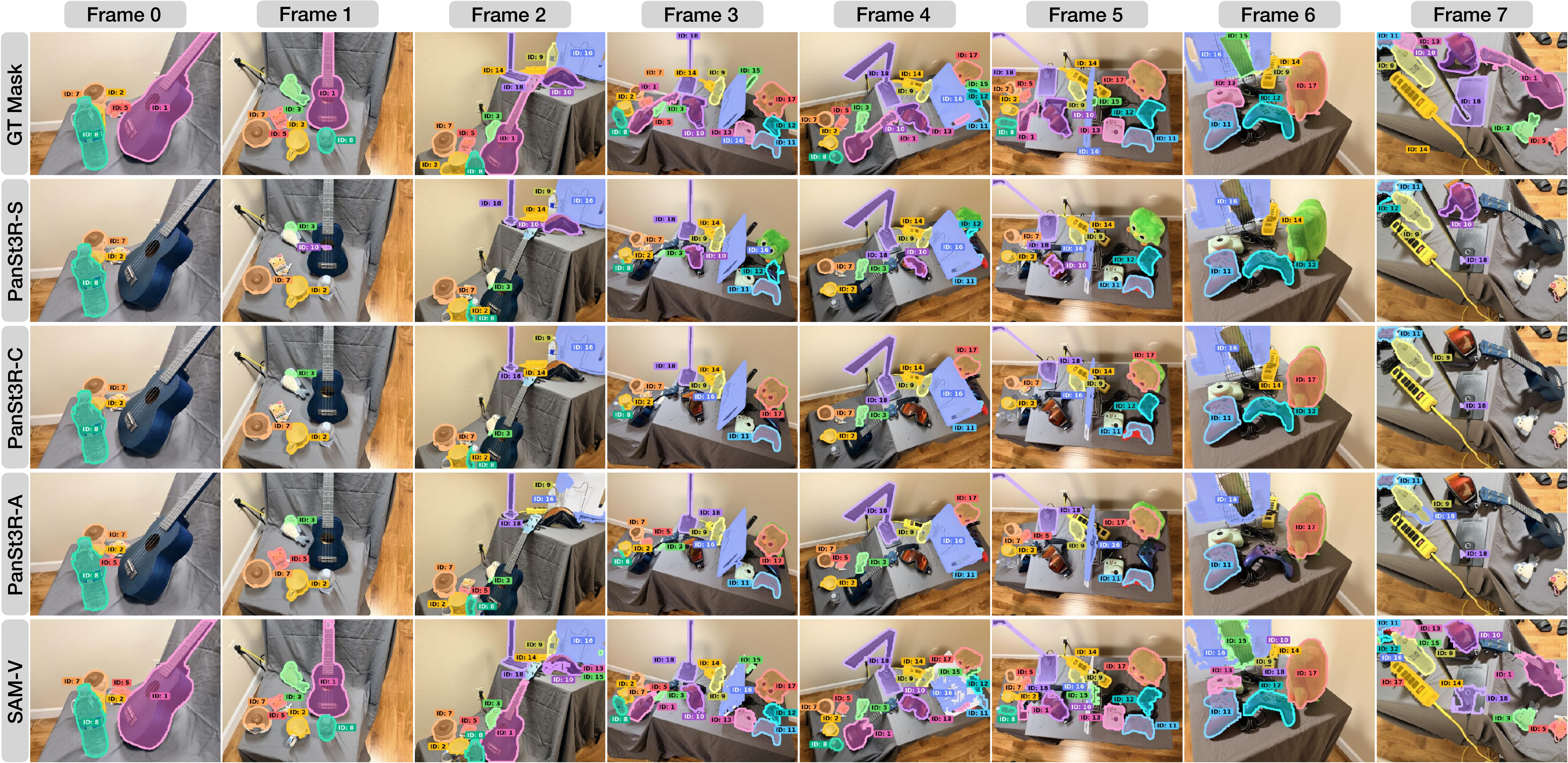}
\caption{\textbf{Multi-view segmentation of a scene containing objects outside
common indoor vocabularies.} Eight views of a tabletop arrangement with sixteen
annotated objects. PanSt3R-S, PanSt3R-C, and PanSt3R-A are the same released
model conditioned on the ScanNet++ ($100$ names), COCO ($133$), and ADE20K
($150$) class lists; PanSt3R-S is the configuration used in Tab.~\ref{tab:iggt_tracking_results} of the main
paper. Colors and IDs denote ground-truth identity, assigned by the Hungarian
matching of Sec.~\ref{sec:supp_metrics}; a track is painted only where its
pooled IoU with the assigned object reaches $0.15$, so an object left unpainted
in a row is one for which that configuration produced no corresponding
instance. The ukulele, camera, and knife are absent from every PanSt3R row and
are recovered by none of the vocabularies, while the plush owl and plush cat
appear and disappear as the class list changes. SAM-V is prompted from points
and segments all sixteen objects without reference to a class list.}
\label{fig:supp_ood}
\end{figure*}

Predicted tracks are matched to ground-truth tracks by the Hungarian assignment
of Sec.~\ref{sec:supp_metrics}. Since the assignment is one-to-one and complete
over the ground truth, it hands every ground-truth object some leftover track
irrespective of overlap; a track is therefore painted in
Fig.~\ref{fig:supp_ood} only when its pooled IoU with the assigned ground-truth
object reaches $0.15$. Objects below that value appear unpainted, which is what
the underlying prediction supports. This threshold governs the rendering alone.
All numbers in Tab.~\ref{tab:supp_ood_coverage} are the raw pooled IoU of the
Hungarian pair, and the metrics elsewhere in the paper use the unmodified
assignment.

\subsection{Results}

Tab.~\ref{tab:supp_ood_coverage} lists every ground-truth object and its pooled
IoU under each configuration. The sixteen objects fall into three groups.

Eight are recovered under every vocabulary. Five are vocabulary-dependent: the
plush cat, the goggles, the purple controller, and the power strip are each lost
under one or two lists, and the plush owl moves from $0.012$ under S to $0.807$
under C and $0.792$ under A, once \texttt{teddy bear} and \texttt{plaything,
toy} enter the class list. The plush owl shows that the mask branch could
segment the object all along, and that the missing element was a name for it.

\begin{table}[htbp]
\centering
\caption{\textbf{Per-object pooled IoU under three class vocabularies and their
union.} Entries marked -- are below the
$0.15$ display threshold and appear unpainted in Fig.~\ref{fig:supp_ood};
the value in parentheses is the pooled IoU of the forced Hungarian pair, which
reflects the size of the segment the object was absorbed into. Category names
are recorded for reference and are read by neither method.}
\label{tab:supp_ood_coverage}
\footnotesize
\setlength{\tabcolsep}{4pt}
\begin{tabular}{@{}lrrrrr@{}}
\toprule
Object & S & C & A & U & SAM-V \tabularnewline
\midrule
mug              & $0.804$ & $0.804$ & $0.815$ & $0.804$ & $0.636$ \tabularnewline
plush seal       & $0.191$ & $0.152$ & $0.191$ & $0.191$ & $0.824$ \tabularnewline
projector        & $0.923$ & $0.892$ & $0.909$ & $0.923$ & $0.850$ \tabularnewline
bottle (low)     & $0.605$ & $0.604$ & $0.619$ & $0.605$ & $0.927$ \tabularnewline
bottle (high)    & $0.729$ & $0.743$ & $0.708$ & $0.729$ & $0.884$ \tabularnewline
controller (red) & $0.887$ & $0.853$ & $0.880$ & $0.891$ & $0.880$ \tabularnewline
cutting board    & $0.874$ & $0.877$ & $0.611$ & $0.874$ & $0.682$ \tabularnewline
lamp             & $0.550$ & $0.498$ & $0.552$ & $0.552$ & $0.679$ \tabularnewline
\midrule
plush cat        & --\,$(0.00)$ & --\,$(0.00)$ & $0.367$ & $0.367$ & $0.608$ \tabularnewline
goggles          & $0.574$ & --\,$(0.00)$ & --\,$(0.03)$ & $0.574$ & $0.680$ \tabularnewline
controller (pur.)& $0.693$ & $0.639$ & --\,$(0.00)$ & $0.693$ & $0.836$ \tabularnewline
power strip      & $0.429$ & $0.423$ & --\,$(0.00)$ & $0.429$ & $0.691$ \tabularnewline
plush owl        & --\,$(0.01)$ & $0.807$ & $0.792$ & $0.792$ & $0.916$ \tabularnewline
\midrule
ukulele          & --\,$(0.00)$ & --\,$(0.15)$ & --\,$(0.00)$ & --\,$(0.00)$ & $0.839$ \tabularnewline
camera           & --\,$(0.01)$ & --\,$(0.00)$ & --\,$(0.00)$ & --\,$(0.01)$ & $0.769$ \tabularnewline
knife            & --\,$(0.00)$ & --\,$(0.00)$ & --\,$(0.00)$ & --\,$(0.00)$ & $0.620$ \tabularnewline
\midrule
recovered        & $11/16$ & $11/16$ & $10/16$ & $13/16$ & $16/16$ \tabularnewline
mean IoU         & $0.455$ & $0.465$ & $0.405$ & $0.527$ & $0.770$ \tabularnewline
\bottomrule
\end{tabular}
\end{table}

For the ukulele, camera, and knife, the absence is at the level of instance
generation. Under S, no predicted track covers $97\%$, $70\%$, and $87\%$ of
their pixels respectively, and the tracks the matcher assigns them are
\texttt{floor} and \texttt{table} segments at IoU $\leq 0.007$. Under C the
ukulele and camera are absorbed into a \texttt{dining table} segment that
contains $93\%$ of each object while consisting of them to $15.1\%$ and $2.1\%$,
so the ukulele's nominal $0.149$ reflects the size of the surrounding segment.
The knife is a useful control: it is named in the COCO list and still scores
$0.001$ there. Vocabulary coverage is necessary for PanSt3R to emit an instance
and does not suffice on its own, since thin and small objects also have to be
captured by one of the scene-level queries. In the same direction, U
behaves as a per-object maximum over the three lists rather than as a function
of their combined length: A is the longest single list and recovers the fewest
objects.

Across the sweep, PanSt3R recovers between $10$ and $13$ of the sixteen objects
depending on which class list it is given, at a mean pooled IoU between $0.405$
and $0.527$. Choosing a vocabulary well is worth $0.12$ mean IoU and three
objects here, and the choice has to be made before the scene is observed. A
prompt-conditioned model carries no such parameter.

\end{document}